\pdfoutput=1

\documentclass[11pt]{article}

\usepackage{ACL2023}

\usepackage{times}
\usepackage{latexsym}
\usepackage{booktabs}
\usepackage{graphicx}
\usepackage{amsmath}

\usepackage[T1]{fontenc}

\usepackage[utf8]{inputenc}

\usepackage{microtype}

\usepackage{inconsolata}

\usepackage{float}

\usepackage[hang,flushmargin]{footmisc} 

\usepackage{enumitem}

\title{NLP-LTU at SemEval-2023 Task 10: The Impact of Data Augmentation and Semi-Supervised Learning Techniques on Text Classification Performance on an Imbalanced Dataset}

\author{Sana Sabah Al-Azzawi \, György Kovács \, Filip Nilsson \, Tosin Adewumi \\ {\bf Marcus Liwicki} 
\\
 EISLAB Machine Learning, Luleå University of Technology, 977 54 Luleå, Sweden \\
 \texttt{firstname.lastname@ltu.se}
\\}

\begin{document}
\maketitle

\begin{abstract}
In this paper, we propose a methodology for  task 10 of SemEval23, focusing on detecting and classifying online sexism in social media posts. The task is tackling a serious issue, as detecting harmful content on social media platforms is crucial for mitigating the harm of these posts on users. Our solution for this task is based on an ensemble of fine-tuned transformer-based models (BERTweet, RoBERTa, and DeBERTa). To alleviate problems related to class imbalance, and to improve the generalization capability of our model, we also experiment with data augmentation and semi-supervised learning. In particular, for data augmentation, we use back-translation, either on all classes, or on the underrepresented classes only. We analyze the impact of these strategies on the overall performance of the pipeline through extensive experiments. while for semi-supervised learning, we found that with a substantial amount of unlabelled, in-domain data available, semi-supervised learning can enhance the performance of certain models. Our proposed method (for which the source code is available on Github\footnote{github.com/SanaNGU/semeval23-task10-sexism-detection-}\footnote{huggingface.co/NLP-LTU/bertweet-large-sexism-detector}) attains an $F1$-score of 0.8613 for sub-taskA, which ranked us 10th  in the competition.


\end{abstract}

\section{Introduction}
Remarkable technological advancements have made it simpler for people from diverse backgrounds to interact through social media using posts and comments written in natural language. These opportunities, however, come with their own challenges. Hateful content on the Internet increased to such levels that manual moderation cannot possibly deal with it \cite{ gongane2022detection}. Thus, precise identification of harmful content on social media is vital for ensuring that such content can be discovered and dealt with, minimizing the risk of victim harm and making online platforms safer and more inclusive.

Detecting online sexism on social media remains a challenge in natural language processing (NLP), and the Explainable Detection of Online Sexism (EDOS) shared task on SemEval23  \cite{kirkSemEval2023} addresses this problem. The task has three main sub-tasks: (i) task A; binary sexism detection, in which we determine whether a given sentence contains sexist content, (ii)  
task B; sexism classification, which places sexist sentences into four categories: threats, derogation, animosity, and prejudiced discussions, and (iii) task C; fine-grained vector of sexism, an eleven-class categorization for sexist posts in which systems must predict one of 11 fine-grained vectors.

One major challenge of  this task is the imbalanced class distribution. 
For instance, sub-task A consists of only 3398 sexist posts, and 10602 non-sexist ones. Using an imbalanced dataset to train models can result in prediction bias towards the majority class \cite{johnson2019survey}.

In this paper, we (team NLP-LTU) present the automatic sexism detection system developed and submitted to SemEval23 task 10; EDOS. The objective of this study is (i) to examine how different state-of-the-art pre-trained language models (PLM) perform in sexism detection and classification tasks,  and (ii) to contribute towards answering  the following research question (RQ): \textbf{ To what extent can data augmentation improve the results and address the data imbalance problem?}


The core of our approach is a voting-based ensemble model consisting of three pre-trained language models: BERTweet-large \cite{nguyen-etal-2020-bertweet}, DeBERTa-v3-large \cite{he2021debertav3}, and RoBERTa-large \cite{liu2019roberta}. Additionally, in order to address the issue of data imbalance and to expand our dataset, our system's pipeline employed techniques such as data augmentation and semi-supervised learning.
We achieved competitive results, ranking us in the top ten for Task A.\footnote{https://github.com/rewire-online/edos/blob/main/leaderboard}
Our results suggest that (i) using PLMs trained on domain-specific data (e.g. BERTweet-large) leads to better results than using PLMs pre-trained on other sources (ii) In most cases extending all classes via augmentation leads to higher classification scores than using augmentation on the minority classes only to completely balance the class distribution.
However, drawing conclusive inferences would require further experiments with multiple data augmentation methods and datasets.
(iii) with a substantial amount of unlabelled, in-domain data available, semi-supervised learning can enhance the performance of certain models.

The rest of the paper is organised as follows: in Section \ref{sec:background}, we present prior related work; in Section \ref{sec:system overview}, we discuss the proposed system. Then, we describe the experiments in Section \ref{sec:experimental setup}. Section \ref{sec:results}, presents results and error analysis. Finally, we conclude the work in Section \ref{sec:conclusion} and describe what further has to be done.

\section{Related Work}
\label{sec:background}

	
	
In the following section we discuss already existing efforts on the detection of sexism, and efforts directed at data augmentation. 

\subsection{Sexism Detection}
Detecting  sexism in social media is essential to ensure a safe online environment and to prevent the negative impact of being a target of sexism. Therefore, several studies have developed datasets and machine-learning models to identify and detect sexism in social media.
Waseem and Hovy's early study involves collecting 16K English tweets and annotating them into three categories: racism, sexism, and neutral \cite{waseem2016hateful}. Similarly, but from a multilingual perspective, \citet{rodriguez2020automatic} created the MeTwo dataset to identify various forms of sexism in Spanish Tweets, and they use machine learning techniques, including both classical and deep learning approaches. Several additional datasets have since been created to examine a wide range of sexist statements \cite{parikh2019multi,samory2021call,rodriguez2021overview,rodriguez2022overview}.
 
The aforementioned studies often categorize sexist content into a limited number of classes, typically two to five, without any further breakdown. However, sexist sentences/posts should be identified, and the reasons for the identification should be provided to increase the interpretability, confidence, and comprehension of the judgments made by the detection system. The EDOS  \cite{kirkSemEval2023} task aims to target this problem with fine-grained classifications for sexist content from social media.

\subsection{Data Augmentation}
A dataset may have several shortcomings that make text classification difficult. This paper mainly focuses on using data augmentation to deal with class imbalance. 
Easy Data Augmentation  (EDA) \cite{wei-zou-2019-eda} use four simple word-based operations to generate new data: synonym replacement, random insertion, random swap, and random deletion. EDA shows that the classification performance  improves even with a simple data augmentation approach. Similarly, \citet{kobayashi2018contextual} stochastically replaces words in the sentences with other relevant words using bidirectional recurrent neural networks. 

\begin{figure*}[t]
    \centering    \includegraphics[width=0.95\textwidth]{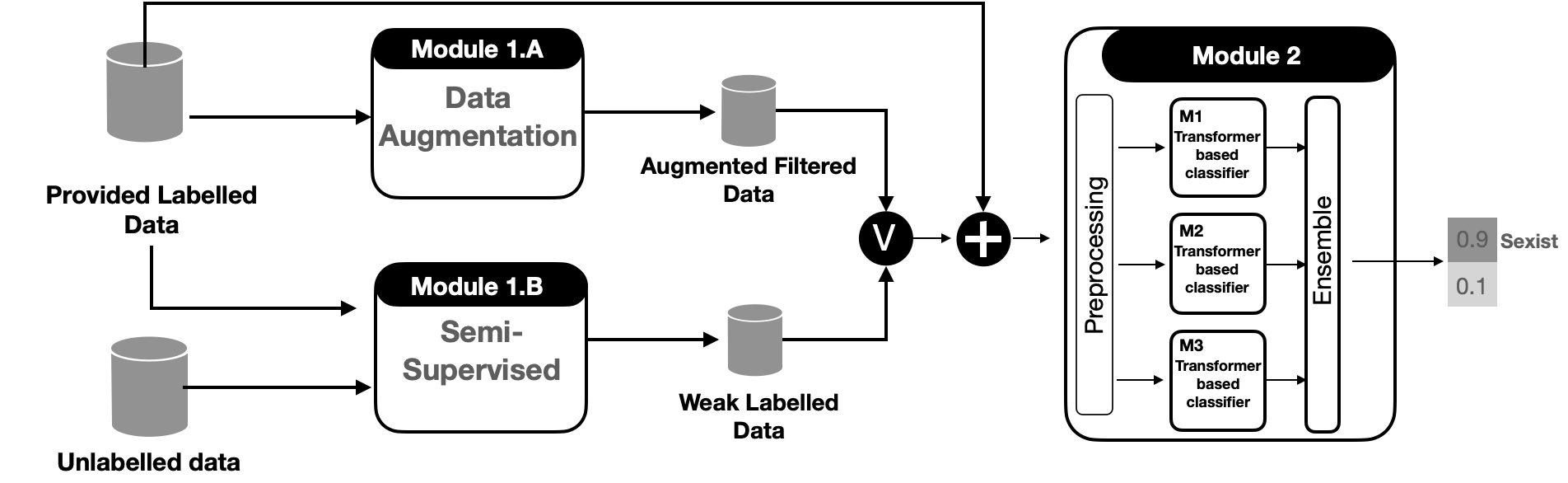}
    \caption{Architecture of the proposed approach}
    \label{fig:overall}
\end{figure*}

In more recent studies, PLM are used to get text samples that are more diverse and linguistically correct.
\citet{anaby2020not} apply GPT-2 to generate synthetic data for a given class in text classification tasks.
Another study by \citet{sabry2022hat5}, uses conversational model checkpoint created by \citet{adewumi2021sm}. 

\section{System Overview}
\label{sec:system overview}
 This section outlines the system pipeline employed in our study, as depicted in Figure \ref{fig:overall}. The proposed approach entails two main stages, generating additional training samples (Module 1), and classification (Module 2). Each is described in its own subsection below. 

%
 

\begin{figure}[b!]
    \centering
    \includegraphics[width=0.5\textwidth]{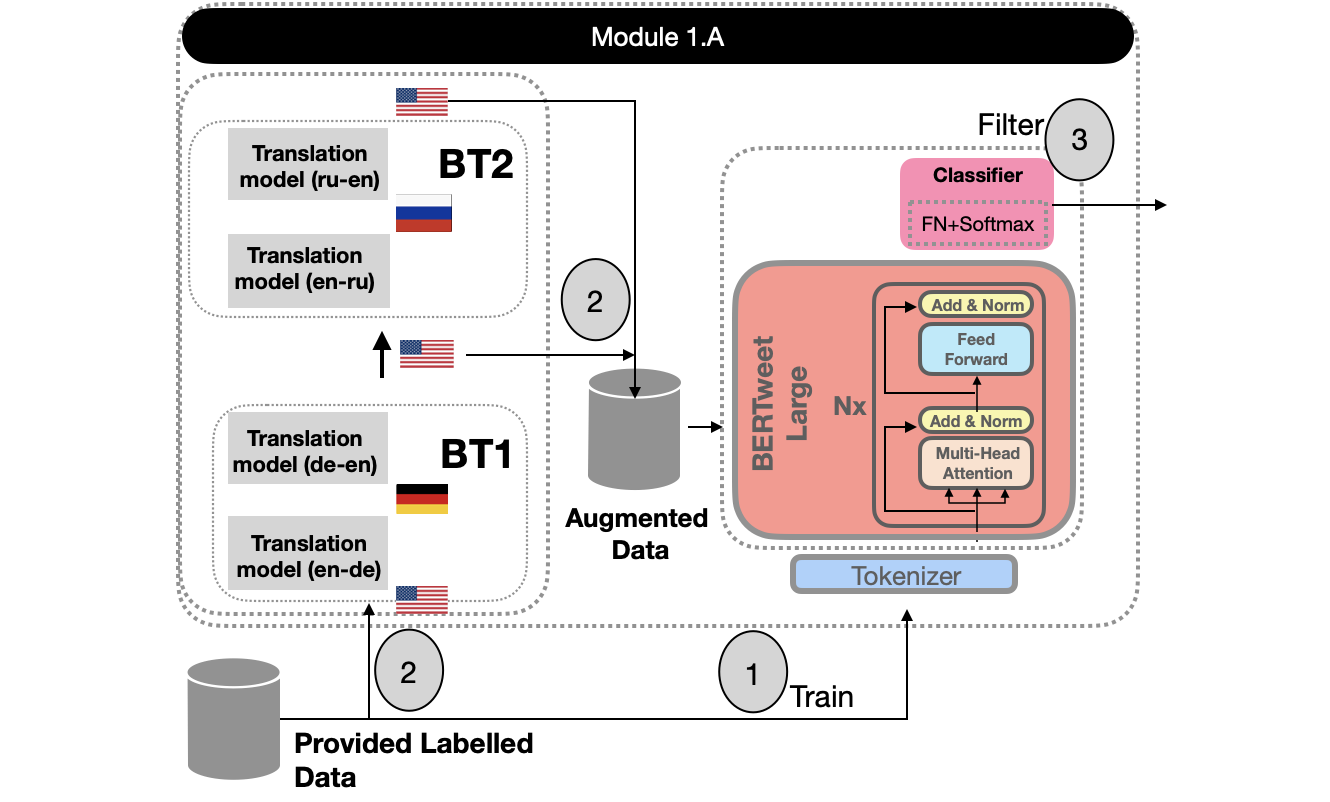}
    \caption{ Back translation data augmentation Block }
    \label{fig:aug}
\end{figure}

\subsection{Module 1.A: Data Augmentation}
Imbalanced data might impede a model's ability to distinguish between highly-represented classes (e.g., non-sexist) and under-represented ones (i.e., sexist). To address this concern, we studied the potential influence of data augmentation approaches on the system's performance.

We expand module 1.A of  Figure \ref{fig:overall} to describe the data augmentation module (shown in Figure \ref{fig:aug}).
The module comprises three main steps. First, we fine-tune our best-performing model; (BERTweet-large) using the gold-labelled data. Then, each sentence  undergoes two rounds of back translation (English to German, and back to English, then English to Russian, and back to English again).
Here, our choice of data augmentation method was motivated by its simplicity and the fact that it does not rely on specific task data and It can be applied independently of the task at hand \cite{longpre2020effective}.


In the final step, the newly generated English sentences from each stage in the second step are filtered using the fine-tuned model from step one. This ensures that each new synthetic sentence retains its original label. This technique can be employed in two ways. Firstly, it can augment only the underrepresented class (sexist sentences to balance the dataset. Alternatively, both classes can be augmented to double the dataset. We investigate the performance of the data augmentation technique using both ways.


\subsection{Module 1.B: Semi-supervised Learning}
Two more unlabelled datasets, each with one million entries, were made available by the task's organizers. 
Inspired by earlier research (e.g.~\cite{shams2014semi}), we used the provided unlabelled datasets to generate weakly labelled samples to balance the original dataset. 

\begin{figure}[b!]
    \centering    \includegraphics[width=0.49\textwidth]{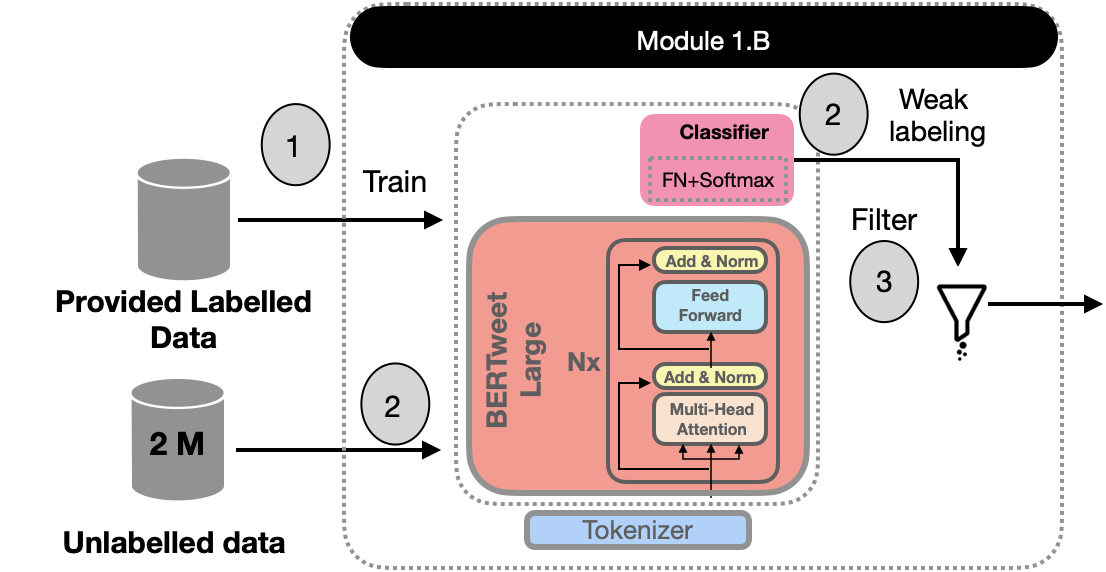}
    \caption{ Semi-Supervised Block.}
    \label{fig:semi}
\end{figure}

As shown in Figure~\ref{fig:semi}, Module 1.B comprises three stages. The first stage being fine-tuning a select pre-trained model (BERTweet-large), using the gold labels. Then, we use the resulting model to create weak labels for the unlabelled data. Lastly, we select samples labelled with a minority class, where the predicted probability of the weak label is at least 0.9.

\subsection{Module 2: Ensemble}
Similar to the full pipeline, Module 2 can also be broken down into its individual constituents, which are (i) the pre-processing module, (ii) the individual classifiers, and (iii) the ensembling method to combine the decision of these classifiers. Firstly, a pre-processing step is needed, as the data for the tasks was collected from noisy resources (Reddit, and Gab).
For this, we used the same common techniques for all models. 
In particular, we converted all uppercase characters to lowercase, removed repetitive patterns like "heeeey" and additional spaces, eliminated special characters like emojis and hashtags (\#), and deleted numbers. 

For the individual classifiers, we examined different PLMs such as BERT \cite{devlin-etal-2019-bert}, RoBERTa  , and DeBERTa. each of these models was initially fine-tuned using the entire dataset.

Lastly, we  employed  an ensemble of the three best-performing classifiers from the previous step for the final submission, namely BERTweet-large, DeBERTa-v3-large, and RoBERTa-large. Ensembling multiple models can potentially prevent egregious mistakes made by a single model \cite{ruta2005classifier,zhang2014weighted}. 
 
We used two ensemble approaches: majority voting, a hard voting method where the prediction of each classifier is treated as a vote, and the class with the most votes is ultimately selected as the predicted class, and soft average ensemble, in which the output of each model is  averaged as shown in  Equation \ref{eq}. 

\begin{equation}\label{eq}
y_{final}={arg max(\frac{y_1+y_2+y_3}{3} )} 
\end{equation}


\section{Experimental Setup}
\label{sec:experimental setup}
All experiments have been implemented using the PyTorch and HuggingFace libraries \cite{wolf2020transformers} on an 8 32GB Nvidia V100 GPU-equipped DGX-1 cluster. The server contains 80 CPU cores with the Ubuntu 18 operating system.
When evaluating solutions, the macro-averaged $F1$-score was the primary metric.
\subsection{Task A : Binary Sexism Detection }
In Task A, we employed the proposed pipeline illustrated in Figure 1. 
Initially, we utilized module 1.A to augment the sexist samples, thereby achieving dataset balance. Subsequently, we integrated the synthetic data with the original data and fine-tuned various pre-trained language models.

The batch size is set to 16, and the Adamw optimizer is used for training. We set the learning rate of the pre-trained model for each language model to 1e-5 and fine-tuned it for three epochs. 
In the semi-supervised learning context, we utilized identical parameters and generated 7,000 additional samples with sexist content to balance the dataset.
\subsection{Task B:sexism classification }
For task B, we excluded Module 1.B. from our pipeline, as in our initial experiments, we were not able to train a sufficiently reliable classifier for the weak-labelling on this smaller dataset. Our hyper-parameters for this task were the same as discussed above, with the exception of an increased number of epochs (4) used here.

About task B, we exclusively employed Module 2 and Module 1.A. Our rationale for this choice stemmed from the inadequacy of the training dataset, which rendered it unfeasible to produce weak labels for this particular task. 

\subsection{Task C : Fine-grained Vector of Sexism}
In our experiments for task C, due to limited time, we forewent the first modules, and focused on Module 2, fine-tuning several pre-trained language models. We have, however, only used these models individually, as the fine-tuned models did not attain comparable levels of performance to those achieved in the previous tasks. Furthermore, the use of an ensemble in such cases may potentially detract from the overall performance of the system.

\begin{table}[b]
\small
\centering
\resizebox{0.9\columnwidth}{!}{
\begin{tabular}{llllll}
\hline
\multicolumn{1}{c}{Model}       & \multicolumn{1}{c}{w/o DA} & \multicolumn{1}{c}{\begin{tabular}[c]{@{}l@{}}with DA-\\ double\end{tabular}}  & \multicolumn{1}{c}{\begin{tabular}[c]{@{}l@{}}with DA-\\ balanced\end{tabular}}  & \multicolumn{1}{c}{\begin{tabular}[c]{@{}l@{}}semi-\\ supervised\\ double-\end{tabular}}&\multicolumn{1}{c}{\begin{tabular}[c]{@{}l@{}}semi-\\ supervised-\\balanced\end{tabular}}    \\\hline

BERT-base               & 82.00      & 81.5  & 78.00& 81.79 &82.10 \\\hline
RoBERTa                 & 83.00     & 83.5   & 81.00& 83.72 & 82.45\\\hline
HateBERT               & 83.58     &  83.00  & 80.01& 84.25 &83.93 \\\hline
RoBERTa-Large            & 84.00   &   84.00 &83.02&85.19 & 85.87 \\\hline
BERTweet-base          & 84.00  &  84.00   & 82.00& 84.73 & 85.68\\\hline
DeBERTa-large-v3       & 86.04    & 84.5   &  83.03&86.39 & 85.47 \\\hline
BERTweet-large       & 86.55        &86.50 & 83.10& 86.07 & 86.12\\\hline 
Soft Ensemble        & 86.73        & 86.01 & 83.08& 86.31& 86.00 \\\hline
Hard Ensemble       & \textbf{86.85} & 86.07 & 83.23& 86.19& 86.01 \\\hline
                 &          
\end{tabular}
}
\caption{\label{table:taska}
$F1$-Macro performance for Task A.
}
\end{table}

\section{Results}
\label{sec:results}
\subsection{Evaluation Phase}
During the evaluation phase, we used the development set provided by the organizers. The results for task A are shown in Table 1. Concerning the data augmentation component, we compared two distinct data augmentation strategies. The initial approach entailed doubling the size of the entire dataset, while the alternative strategy solely augmented samples that contained sexist content.

\begin{table}[t]
\small
\centering
\begin{tabular}{lll}
\hline
\multicolumn{1}{c}{Model}                                           & \multicolumn{1}{c}{w/o DA}& \multicolumn{1}{c}{with DA} \\ \hline
BERT-base      & 60.33   & 62.33                     \\ \hline
BERTweet-base    &56.33     & 59.05                      \\ \hline
BERT-large                  & 60.66  & 63.66                      \\ \hline
RoBERTa                    & 59.33 & 59.33                      \\ \hline
RoBERTa-Large              & 68.00 & 68.33                       \\ \hline
DeBERTa-large-v3        & 68.33  & 67.16                     \\ \hline
BERTweet-large             & 67.33 & 66.00                      \\ \hline
Soft Ensemble                                                  & 69.99 & 69.00                       \\ \hline
Hard Ensemble                                                 & \textbf{70.30} & \textbf{70.00}              \\ \hline
\end{tabular}
\caption{\label{table:taskb}
$F1$-Macro performance for Task B.
}
\end{table}
Table \ref{table:taskb} shows the results on the development set for Task B. Due to the limited size of the training set, the use of data augmentation techniques resulted in an improved performance for some models, while others exhibited similar $F1$-scores to those obtained without augmentation.
\begin{table}[b]
\small
\centering
\begin{tabular}{lll}
\hline
\multicolumn{1}{c}{Model} & \multicolumn{1}{c}{w/o DA} & \multicolumn{1}{c}{with DA} \\ \hline
BERTweet-base             & 30.01 &   30.33                   \\ \hline
BERT-base                 & 30.00   & 30.66                     \\ \hline
RoBERTa-base              & 34.01    & 36.66                     \\ \hline
DeBERTa-large                & 38.33  &   38.66                   \\ \hline
BERT-large                & 42.33   &    41.66                 \\ \hline
BERTweet-large            & 45.66   &   45.33                  \\ \hline
RoBERTa-large             & \textbf{47.33}    &  46.66                 \\ \hline
\end{tabular}
\caption{\label{table:taskc}
$F1$-Macro performance for Task C.
}
\end{table}

The results shown in Table 1 indicate that the use of the provided dataset with a hard ensemble strategy yields the best performance. Furthermore, the semi-supervised approach improves the performance of some pre-trained models (e.g. BERT-base, HateBERT \cite{caselli2020hatebert}, BERTweet-base), but not those models, which had been 
pre-trained on larger datasets (e.g. DeBERTa-large-v3, BERTweet-large). We hypothesize that these larger models already possess more knowledge due to their extensive pre-training. Regarding data augmentation, our findings indicate that doubling all classes resulted in better performance than balancing the dataset.

\subsection{Test Phase}
We combined the training and development data during the test phase and fine-tuned the models. Our submission, as demonstrated in Table 1, was only made once. We utilized only Module 2 from our pipeline for Task A, employing the hard ensemble strategy. Our three top-performing models, BERTweet-large, DeBERTa-v3-large, and RoBERTa-large, were used without data augmentation.
The same approach was adopted for Task B. For Task C, we used RoBERTa-large for the final submission, which yielded the best results in the evaluation set.
\begin{table}[t!]
\small
\centering
\caption{\label{table:test}
Results on the Test set for All Tasks}
\begin{tabular}{llll}
\hline
Task & Model         & F1-score  &  Rank \\ \hline
A    & Ensemble      & 86.13   &  10 \\ \hline
    B    & Ensemble      & 65.50    & 18    \\ \hline
C    & RoBERTa-large & 46.00      &  23 \\ \hline
\end{tabular}
\end{table}
\subsection{Error Analysis}
\begin{figure}[h!]
\centering
\includegraphics[width=0.5\textwidth]{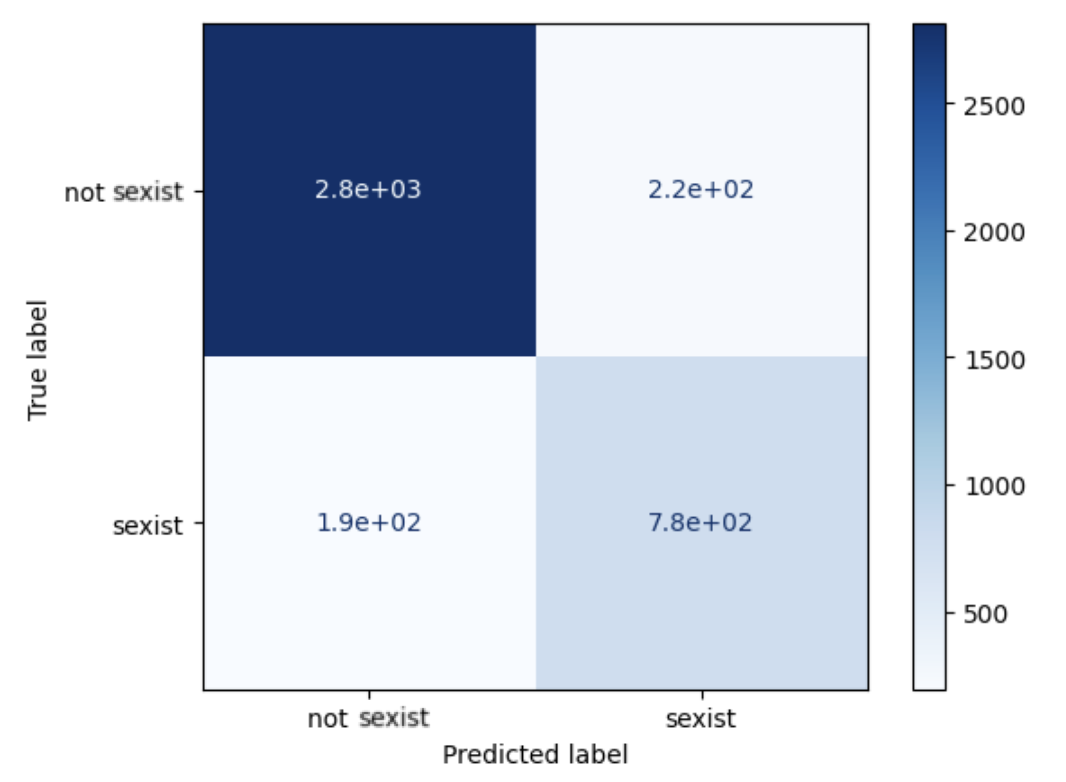}
\caption{Confusion Matrix for  TaskA}
\label{CM}
\end{figure}

In this subsection, we have undertaken an error analysis for the submission on Task A. The confusion matrices presented in Figure \ref{CM} was%
constructed to evaluate the performance of our models on the test set. Our ensemble model achieved an $F1$-score of 86.13 on the test set for task A.  However, the confusion matrix illustrated in Figure \ref{CM} indicates that the model correctly predicted the (not sexist) class 92.40 \% of the time (2,800 out of 3,030), while struggling to generate correct predictions for the (sexist) class, with a correct prediction rate of only 80.41\% (780 out of 970).

This discrepancy is most likely due to the data imbalance, as 85.7\% of the total training set comprises samples labelled as (not sexist). Despite performing data augmentation using back-translation to mitigate the data imbalance issue, the results in the Table\ref{table:taska} indicate that this technique did not improve the overall performance. We hypothesise that the back-translation method did not generate diverse samples, and one possible solution is to use data augmentation methods that generate more diverse synthetic data.



\section{Conclusion}
\label{sec:conclusion}

This paper presents our solution to the Shared Task on Explainable Detection of Online Sexism at SemEval23. Our approach involved employing ensemble voting techniques with previously fine-tuned language models, specifically BERTweet-large, RoBERTa-large, and DeBERTa-V3-large, which resulted in the best performance for both task A and B. Additionally, we discovered that fine-tuning RoBERTa-Large was the most effective approach for addressing task C, outperforming the ensemble voting method. These findings address the first objective of examining how different
state-of-the-art transformer-based models perform
in sexism detection and classification.


To address our  research question; \textbf{(RQ): to what extent can data augmentation improve the results and address the data imbalance problem}, we employed a task agnostic data augmentation method, specifically back-translation, in two scenarios: one to double the dataset and the other to augment the underrepresented class. Our results showed that augmenting all classes was more effective than balancing the dataset by augmenting only the underrepresented class, which motivates further exploration of the effects of data augmentation on text classification with unbalanced datasets.
In future research, we plan to explore alternative data augmentation techniques to produce more diverse sentences, such as utilizing generative models like GPT-2, to balance and double the dataset's size, and compare the results with the back-translation method.

Moreover, we plan to investigate  why augmenting all classes sometimes was more effective  than augmenting only the underrepresented class and  balancing the dataset.
\section*{Acknowledgements}
This work was partially supported by the Wallenberg AI, Autonomous Systems and Software Program (WASP) funded by the Knut and Alice Wallenberg Foundation.

\bibliography{custom}
\bibliographystyle{acl_natbib}

\end{document}